\documentclass{article}

\usepackage{arxiv}

\usepackage[utf8]{inputenc} 
\usepackage[T1]{fontenc}    
\usepackage{hyperref}       
\usepackage{url}            
\usepackage{booktabs}       
\usepackage{amsfonts}       
\usepackage{nicefrac}       
\usepackage{microtype}      
\usepackage{lipsum}		
\usepackage{graphicx}
\usepackage{natbib}
\usepackage{doi}
\usepackage{amsmath}

\title{Enhancing Fetal Plane Classification Accuracy with Data Augmentation Using Diffusion Models}

\author{{Yueying Tian} \\
	School of Engineering \\and Informatics\\
	University of Sussex\\
	Brighton, United Kingdom\\
	\And
	{Elif Ucurum} \\
	School of Engineering\\ and Informatics\\
	University of Sussex\\
	Brighton, United Kingdom \\
    \And
	{Xudong Han} \\
	School of Engineering\\ and Informatics\\
	University of Sussex\\
	Brighton, United Kingdom \\
	\AND
	{Rupert Young} \\
	School of Engineering\\ and Informatics\\
	University of Sussex\\
	Brighton, United Kingdom \\
    \And
    {Chris Chatwin} \\
	School of Engineering\\ and Informatics\\
	University of Sussex\\
	Brighton, United Kingdom \\
    \And
    {Philip Birch}\thanks{Corresponding author} \\
	School of Engineering\\and Informatics\\
	University of Sussex\\
	Brighton, United Kingdom \\
    \texttt{P.M.Birch@sussex.ac.uk}\\
}

\date{}

\renewcommand{\headeright}{}
\renewcommand{\undertitle}{}
\renewcommand{\shorttitle}{Enhancing Fetal Plane Classification Accuracy with Data Augmentation Using Diffusion Models}

\hypersetup{
pdftitle={A template for the arxiv style},
pdfsubject={q-bio.NC, q-bio.QM},
pdfauthor={David S.~Hippocampus, Elias D.~Striatum},
pdfkeywords={First keyword, Second keyword, More},
}

\usepackage{subcaption}
\usepackage[justification=centering]{caption}

\begin{document}
\maketitle

\begin{abstract}
Ultrasound imaging is widely used in medical diagnosis, especially for fetal health assessment. However, the availability of high-quality annotated ultrasound images is limited, which restricts the training of machine learning models. In this paper, we investigate the use of diffusion models to generate synthetic ultrasound images to improve the performance on fetal plane classification. We train different classifiers first on synthetic images and then fine-tune them with real images. Extensive experimental results demonstrate that incorporating generated images into training pipelines leads to better classification accuracy than training with real images alone. The findings suggest that generating synthetic data using diffusion models can be a valuable tool in overcoming the challenges of data scarcity in ultrasound medical imaging.
\end{abstract}


\renewcommand\thefootnote{}

\renewcommand\thefootnote{\fnsymbol{footnote}}
\setcounter{footnote}{1}

\section{INTRODUCTION}
\label{sec1}

Over the past decade, deep learning has achieved great success in medical image diagnostics, including tasks such as nodule detection \citep{pehrson2019automatic}, image segmentation \citep{wang2022medical} and classification \citep{cai2020review}. Deep learning models, such as convolutional neural networks (CNNs) \citep{lecun1989handwritten}, and residual networks (ResNets) \citep{he2016deep}, trained on large datasets, can learn intricate patterns in medical images, improving diagnostic accuracy and supporting healthcare professionals. However, the successful application of machine learning to medical imaging is heavily dependent on the availability of large, high-quality annotated datasets \citep{alzubaidi2021novel}. In many fields of medicine, obtaining such datasets can be difficult, and this issue is especially pronounced in ultrasound imaging \citep{huang2021dense}.

Ultrasound imaging plays a crucial role in prenatal care, providing valuable and non-invasive insights into fetal health and development \citep{brattain2018machine}.  Despite their widespread use, one of the key challenges in developing deep learning models for ultrasound imaging is data scarcity. There are several reasons why obtaining a large dataset of labeled ultrasound images is difficult.

Firstly, due to privacy concerns, hospitals and medical institutions are often unable to release patient data, making it challenging to build large public datasets \citep{dumont2021overcoming}. Furthermore, even if hospitals are willing to share data, obtaining labeled data is costly and time-consuming. Expert knowledge is required to correctly annotate medical images, and for ultrasound, this usually means that radiologists or sonographers must manually label features of interest, such as fetal planes or abnormalities. This process is not only slow but also expensive, which limits the amount of labeled ultrasound data available for training machine learning models.

Given these challenges, there has been growing interest in the use of synthetic data and generative models to supplement real-world data. Recent advances in generative models, such as Generative Adversarial Networks (GANs) \citep{goodfellow2014generative} and Diffusion Models \citep{ho2020denoising}, have opened up new possibilities for creating realistic synthetic medical images. These models can learn complex data distributions from real images and generate new synthetic images that resemble the originals. This approach has shown promise in a variety of medical imaging applications. For instance, \citet{frid2018gan} demonstrated the use of GANs for generating synthetic liver lesions in CT images. \citet{motamed2021data} showed the effectiveness of using GANs to generate synthetic chest X-ray images for training classifiers. \citet{yi2019generative} provided a comprehensive review of GAN-based techniques for generating synthetic medical images. \citet{pan20232d} demonstrated that diffusion models generate more realistic synthetic CT and MRI images for medical image segmentation than GANs. \citet{hardy2023improving} showed that generating realistic synthetic liver ultrasound images with a diffusion model can enhance liver disease classification performance. More recently, \citet{qiu2025noise} proposed a noise-consistent Siamese-Diffusion model for medical image synthesis and segmentation, further expanding the capabilities of diffusion models in this domain. However, the application of generative models to ultrasound imaging still remains less explored, particularly for fetal ultrasound, where data scarcity is a significant challenge.

In this work, we propose the use of diffusion models to generate synthetic fetal ultrasound images and investigate their potential to improve the performance of machine learning classifiers. Unlike GANs, which can suffer from training instability and mode collapse \citep{kodali2017convergence}, diffusion models offer more stable training dynamics and produce high-quality, diverse samples, making them well-suited for this application \citep{dhariwal2021diffusion}. Specifically, we hypothesize that combining real ultrasound images with generated images can enhance classifier accuracy, especially in cases where annotated real data is limited. By training the model first on synthetic images and then fine-tuning it on real images, we aim to overcome the data scarcity problem in fetal ultrasound imaging. We make the following key contributions:

\begin{itemize}
    \item We generate in total $60000$ synthetic ultrasound images of fetal planes using a trained diffusion model. These images, which will be made publicly available, can serve as a valuable resource for future research in medical image classification.
    \item We conduct experiments with six different classifiers and show that pretraining the classifiers with synthetic images followed by fine-tuning on real images results in improved classification accuracy compared to training only with real images.
    \item To gain deeper insights into the classifier's behavior, we analyze the confusion matrices. Our findings reveal that the improvement in classification performance is particularly significant for imbalanced classes with small numbers of images. 
    \item Furthermore, we observe a consistent trend: as the number of generated synthetic images increases, the classification performance after pretraining and fine-tuning consistently improves. These results demonstrate the effectiveness of leveraging synthetic data to enhance the robustness and accuracy of medical image classifiers, especially in scenarios where real-world data is limited or imbalanced.
\end{itemize}

By addressing the data scarcity issue and demonstrating the potential of synthetic ultrasound images in improving classifier performance, our work opens up new possibilities for the use of generative models in medical image analysis, particularly in fields where obtaining large labeled datasets is challenging.

\section{RELATED WORK}
\label{sec2}

The challenge of data scarcity in medical imaging has led researchers to explore a variety of solutions to augment training datasets. These approaches can broadly be categorized into traditional methods, and the more recent deep learning-based methods. This section provides an overview of the key works in each of these categories.

\subsection{Traditional Approaches} 

Traditionally, data scarcity in medical imaging has been addressed through various data augmentation techniques. These methods aim to artificially expand the size of a dataset without requiring additional labeled images. Early data augmentation strategies for medical images included transformations such as rotation, scaling, cropping, flipping, and elastic deformations \citep{hussain2017differential}. These techniques were particularly useful in fields like MRI and CT imaging, where the generation of new images from the same subject could help the model generalize better \citep{goceri2023medical}. For general imaging, methods such as image cropping and noise addition have been widely used. Simple geometric transformations, such as rotation and scaling have been applied broadly as well. While these methods can be effective in increasing the volume of data, they often fail to generate truly novel variations of the data, which limits their ability to capture the full diversity of possible real-world cases \citep{goceri2023medical}.

Another traditional approach to tackling data scarcity is transfer learning, which has been used in many medical imaging tasks \citep{sanford2020data}. Transfer learning allows models trained on large, publicly available datasets (such as ImageNet) to be adapted to medical datasets with fewer labeled samples. While effective in some contexts, transfer learning still relies heavily on the availability of a reasonable amount of high-quality annotated data, making it less suitable in situations where data is extremely limited or highly specialized, such as in fetal ultrasound imaging.

\subsection{Deep Learning-Based Approaches} 

The rise of deep learning has revolutionized medical image analysis, neural networks becoming the most widely used in medical imaging. However, these methods often require large amounts of labeled data to perform well. This has led to significant interest in using deep learning to generate synthetic medical images as a way of alleviating data scarcity.

In particular, Generative Adversarial Networks (GANs) have been explored as a way to generate synthetic medical images. GANs consist of two neural networks: a generator that creates synthetic data, and a discriminator that attempts to distinguish between real and fake data. GANs have been applied to various imaging modalities, including CT, MRI, and ultrasound. For instance, \citet{frid2018gan} used GANs to generate synthetic CT images of liver lesions, which were used to augment a classifier for liver lesion detection. Similarly, \citet{motamed2021data} showed the effectiveness of using GANs to generate synthetic chest X-ray images for training classifiers. In ultrasound imaging, GANs have shown promise in generating synthetic images for various purposes. \citet{montero2021generative} applied a GAN-based approach to generate synthetic fetal ultrasound images, focusing on augmenting data for training classifiers on fetal development stages. While GANs have shown considerable success, one challenge with these models is the difficulty in training them, especially in medical imaging, where the distribution of data can be complex and diverse. In addition to GANs, Variational Autoencoders (VAEs) have also been used for generating synthetic medical images. VAEs are another class of generative models that learn a probabilistic mapping between a low-dimensional latent space and high-dimensional image space. Studies such as \citet{dorent2023unified} have demonstrated the potential of VAEs for generating synthetic ultrasound and MRI images for the purposes of data augmentation.

In recent years, diffusion models have emerged as a powerful new class of generative models, particularly in the field of image synthesis \citep{ho2020denoising}. Diffusion models work by simulating a gradual ``noising'' process that adds random noise to data in a series of steps, and then by learning how to reverse this process they can generate high-quality samples. One key advantage of diffusion models over GANs and VAEs is their ability to produce high-fidelity images with more stable training dynamics.

The use of diffusion models in medical imaging is a relatively recent development, but promising results have already been reported in various applications. \citet{dhariwal2021diffusion} demonstrated the use of diffusion models for generating high-quality images in a variety of domains, including natural images, and showed that they outperform GANs in terms of image quality and diversity. In medical imaging, recent studies have started to explore diffusion models for generating synthetic data. \citet{pan20232d} applied diffusion models to generate synthetic CT and MRI images for medical image segmentation tasks, showing that diffusion models can generate more realistic images compared to traditional methods like GANs. Similarly, in the field of ultrasound imaging, \citet{hardy2023improving} used a diffusion model to generate realistic synthetic liver ultrasound images, demonstrating that such generated data could improve the performance of a liver disease classification model. However, the application of diffusion models to fetal ultrasound images remains an underexplored area, and there is a gap in the literature regarding how these models can be applied to enhance classification performance for fetal health assessment.

\subsection{Fetal Plane Classification}
\label{sec:existing_work_fetal_plane_classification}
Recent advancements in fetal ultrasound classification have explored diverse approaches. For instance, \citet{krishna2023automated} utilized a multi-layer perceptron with deep feature integration to automate the classification of maternal fetal ultrasound planes. This work demonstrated the potential of integrating deep features for improved classification accuracy. Similarly, \citet{krishna2024standard} employed a stacked ensemble of deep learning models, achieving high performance through the combination of multiple classifiers. However, our method focuses on the performance of a single network enhanced by diffusion models. Also, \citet{rauf2023automated} proposed a deep bottleneck residual 82-layered architecture with Bayesian optimization for ultrasound plane classification. This method differs from our work by its focus on architectural optimization and Bayesian methods, whereas we focus on data augmentation through diffusion models. Transfer learning has also been explored, as seen in \citet{ghabri2023transfer}, which focused on accurate fetal organ classification, showing the potential of pre-trained models in this domain. Furthermore, the works of \citet{krishna2024automatic,krishna2024deep} specifically address the automated identification of fetal biometry planes, utilizing techniques such as adaptive channel weighting. These works highlight the importance of accurate biometry plane identification, which is a key step in fetal assessment. However, they do not explore the use of diffusion models for data augmentation, as we do in our work.

\section{METHODS}
\label{sec3}

In this section, we describe the methodology used to generate synthetic fetal ultrasound images and train a classifier to evaluate their utility in improving classification performance. We employ a classifier-guided diffusion model, which combines the power of generative models and supervised learning to generate class-specific images. Our method involves two main components: the training of the diffusion model and the classification task.

\subsection{Basics of Diffusion Models}

Diffusion models are a class of generative models that have recently shown remarkable success in synthesizing high-quality images. Unlike Generative Adversarial Networks (GANs), which learn to generate images through an adversarial process, diffusion models operate by progressively adding noise to data and then learning to reverse this noisy process to generate new data. The core idea is inspired by non-equilibrium thermodynamics.

\textbf{The Forward (Diffusion) Process.}
In the forward process, a data sample $x_0$ (e.g., an image) is gradually corrupted by adding Gaussian noise over a series of $T$ timesteps. At each timestep $t$, a small amount of noise is added to the previous timestep's sample $x_{t-1}$ to produce $x_t$. This process is defined such that the distribution $q(x_t|x_{t-1})$ is a Gaussian, and consequently, the distribution $q(x_t|x_0)$ can be directly sampled:

$$q(x_t|x_0) = \mathcal{N}(x_t; \sqrt{\alpha_t} x_0, (1-\alpha_t)I)$$

where $\alpha_t$ is a predefined variance schedule, typically decreasing over time, such that $x_T$ eventually becomes pure Gaussian noise.

\textbf{The Reverse (Denoising) Process.}
The goal of a diffusion model is to learn the reverse process, which involves denoising a noisy sample $x_t$ back to a cleaner $x_{t-1}$, and eventually to the original data $x_0$. This reverse process is also modeled as a Gaussian, but with learned means and variances:

$$p_\theta(x_{t-1}|x_t) = \mathcal{N}(x_{t-1}; \mu_\theta(x_t, t), \Sigma_\theta(x_t, t))$$

where $\mu_\theta$ and $\Sigma_\theta$ are learned parameters of a neural network.

\textbf{Training Objective.} \citet{dhariwal2021diffusion} builds upon the denoising diffusion probabilistic models (DDPMs) framework, where the training objective for diffusion models is to maximize the likelihood of the training data. This is achieved by optimizing a variational lower bound (VLB) on the negative log-likelihood.

A simplified and commonly used training objective, particularly in the context of the mentioned paper, is to train a neural network (often denoted as $\epsilon_\theta$) to predict the noise that was added at each timestep $t$. Specifically, the objective function is a mean squared error between the predicted noise and the actual noise.

For a given noisy sample $x_t$ at timestep $t$, obtained by adding noise $\epsilon \sim \mathcal{N}(0, I)$ to $x_0$:

$$x_t = \sqrt{\alpha_t} x_0 + \sqrt{1-\alpha_t} \epsilon$$

The neural network $\epsilon_\theta(x_t, t)$ is trained to predict $\epsilon$. The simplified training objective can be expressed as:

$$\mathcal{L}_{\text{simple}} = \mathbb{E}_{t \sim [1, T], x_0 \sim q(x_0), \epsilon \sim \mathcal{N}(0, I)} \left[ \left\|\epsilon - \epsilon_\theta(x_t, t) \right\|^2 \right]$$

where:
\begin{itemize}
    \item $t$ is sampled uniformly from $1$ to $T$.
    \item $x_0$ is a data sample from the true data distribution $q(x_0)$.
    \item $\epsilon$ is the pure Gaussian noise sampled from $\mathcal{N}(0, I)$.
    \item $x_t$ is the noisy sample at timestep $t$, obtained using the forward process formula above.
    \item $\epsilon_\theta(x_t, t)$ is the neural network's prediction of the noise, conditioned on $x_t$ and the timestep $t$.
    \item $||\cdot||^2$ denotes the squared L2 norm.
\end{itemize}

This objective aims to minimize the difference between the true noise $\epsilon$ and the noise predicted by the model $\epsilon_\theta$. By minimizing this objective, the model learns to accurately predict the noise component at each timestep, which in turn allows it to effectively reverse the diffusion process and generate high-quality samples.

\subsection{Classifier-Guided Diffusion Model}

We use a classifier-guided diffusion model to generate synthetic ultrasound images \citep{dhariwal2021diffusion}. In this approach, the diffusion model is conditioned on the classifier’s predictions, allowing it to generate images that belong to specific classes. This method enhances the model's ability to produce class-specific images that better align with real-world data, improving the quality and relevance of the generated samples. The overall process consists of two interconnected models: the diffusion model and the classifier.

The diffusion model is trained to learn the data distribution of fetal ultrasound images across different classes. A key feature of this approach is that we condition the image generation process on the class labels. Unlike traditional diffusion models that generate images from random noise, the classifier-guided diffusion model is trained such that it can guide the image generation towards specific class categories during the denoising process. During training, the classifier provides the model with class-specific information, which helps the diffusion model generate images that are representative of the corresponding class.

The classifier is trained simultaneously with the diffusion model to predict the class label of a given image. The classifier is a half U-Net (containing only the encoder part) that learns to differentiate between the various classes of fetal ultrasound images. Using a half U-Net allows for efficient feature extraction with fewer parameters, while still capturing essential spatial hierarchies needed for accurate class prediction.

During the training phase, the classifier is used to provide feedback to the diffusion model, ensuring that the generated images match the intended class distribution. The classifier-guided diffusion model allows us to generate high-quality synthetic images that are not only realistic but also tailored to each class of interest, improving the relevance of the synthetic data for training downstream classifiers.

\subsection{Ultrasound Image Dataset} 

To train and evaluate our proposed method, we use the FETAL\_PLANES\_DB \citep{burgos2020evaluation}, a commonly used maternal-fetal ultrasound image dataset. The FETAL\_PLANES\_DB contains in total $12400$ images of fetal planes, which are essential in prenatal care for monitoring fetal development. The dataset is divided into six distinct classes, i.e., Abdomen, Brain, Femur, Thorax, the mother’s cervix (widely used for prematurity screening), and a general category to include any other less common image plane. These classes represent different features that are crucial for assessing fetal health.

\begin{table}[h]
\centering
\begin{tabular}{ c | c | c | c | c } 
\toprule
Class index & Class name & \# images  & \# training images & \# test images \\
\midrule
0 & Fetal abdomen & $711$ & $569$ & $142$ \\
\midrule
1 & Fetal brain & $3092$ & $2474$ & $618$ \\ 
\midrule
2 & Fetal femur & $1040$ & $833$ & $207$ \\
\midrule
3 & Fetal thorax  & $1718$ & $1375$ & $343$ \\ 
\midrule
4 & Maternal cervix  & $1626$ & $1301$ & $325$ \\
\midrule
5 & Other  & $4213$ & $3371$ & $842$ \\
\midrule
\midrule
& Total & $12400$ & $9923$ & $2477$ \\
\bottomrule
\end{tabular}
\caption{FETAL\_PLANES\_DB dataset.}
\label{table:fetal_planes_db_dataset}
\end{table}
The FETAL\_PLANES\_DB dataset provides a valuable resource for training models in fetal ultrasound image classification. However, due to the relatively small size of the dataset, we face the common challenge of data scarcity, which makes it difficult to train robust deep learning models. Additionally, as shown in Table~\ref{table:fetal_planes_db_dataset} there is an imbalance between different classes, which can further hinder model performance. To address these issues, we augment the dataset using the classifier-guided diffusion model.

\begin{figure}[h]
    \centering 
\begin{subfigure}{0.3\textwidth}
  \includegraphics[width=2.1in,height=1.4in]{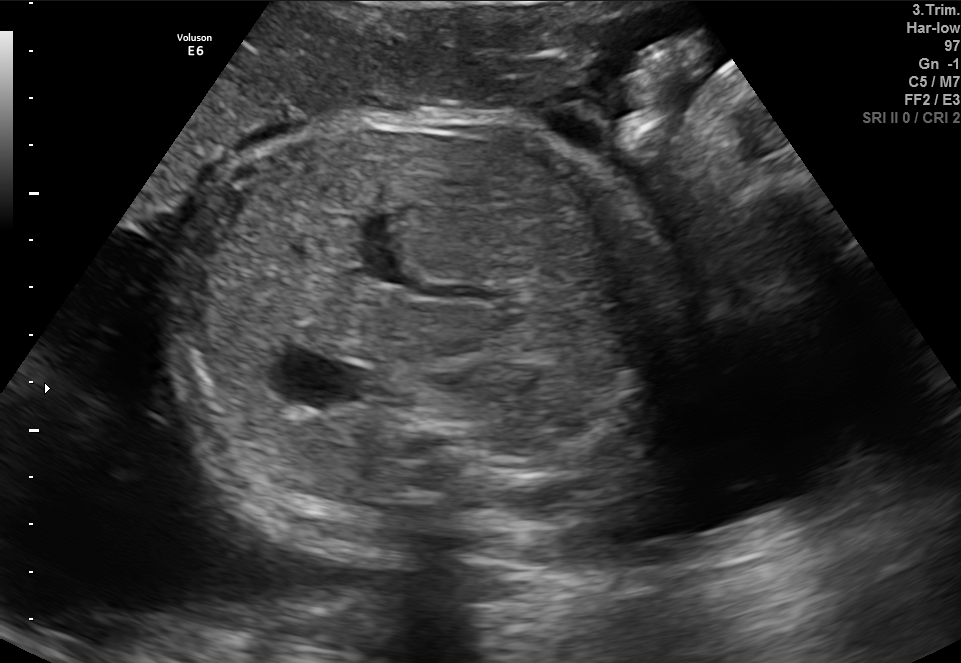}
  \caption{Fetal abdomen}
\end{subfigure}\hfil 
\begin{subfigure}{0.3\textwidth}
  \includegraphics[width=2.1in,height=1.4in]{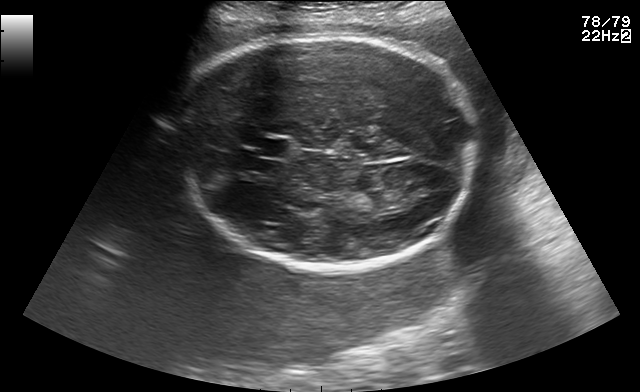}
  \caption{Fetal brain}
\end{subfigure}\hfil 
\begin{subfigure}{0.3\textwidth}
  \includegraphics[width=2.1in,height=1.4in]{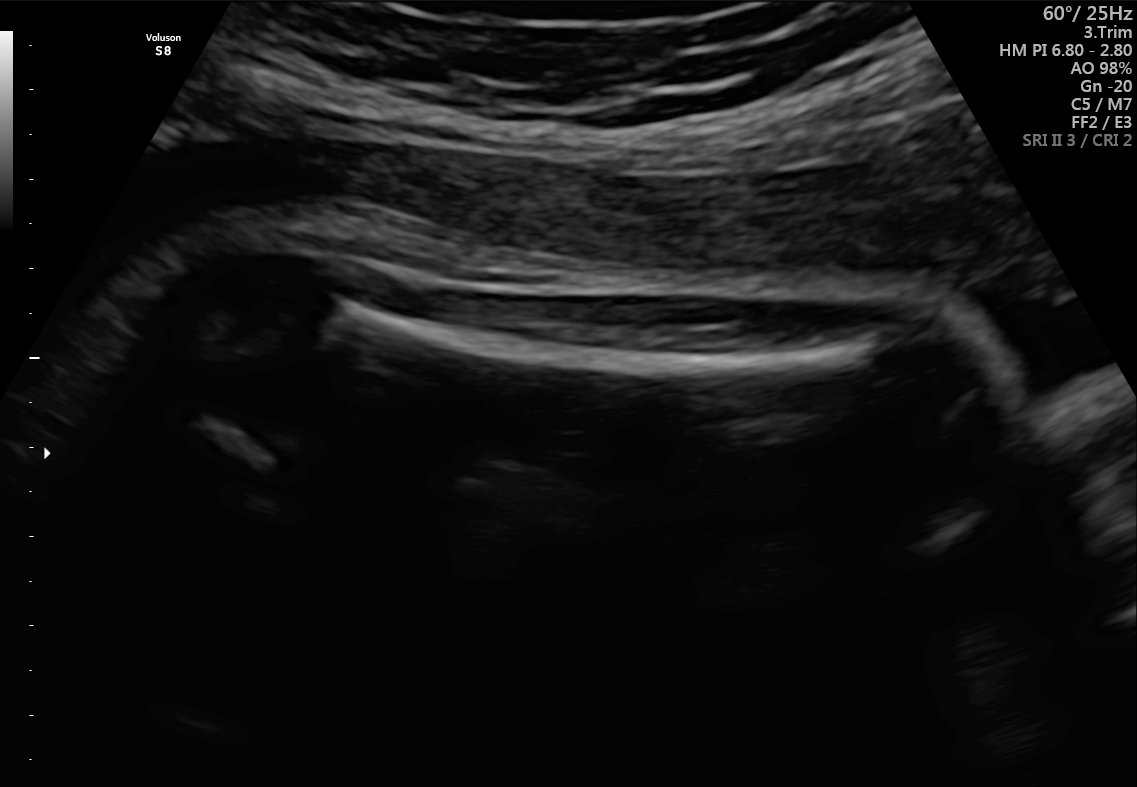}
  \caption{Fetal femur}
\end{subfigure}

\medskip
\begin{subfigure}{0.3\textwidth}
  \includegraphics[width=2.1in,height=1.4in]{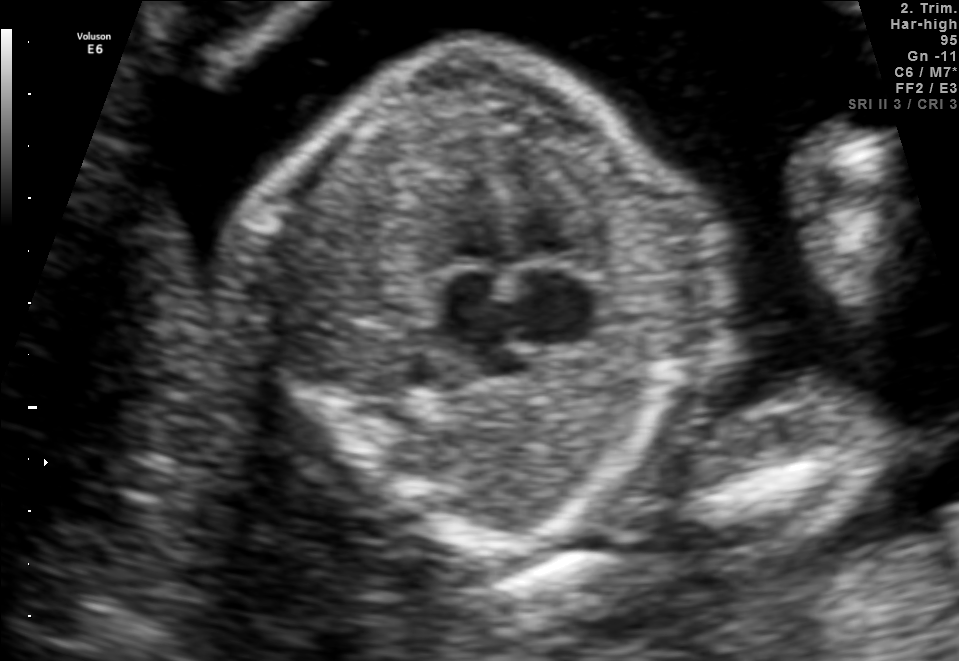}
  \caption{Fetal thorax}
\end{subfigure}\hfil 
\begin{subfigure}{0.3\textwidth}
  \includegraphics[width=2.1in,height=1.4in]{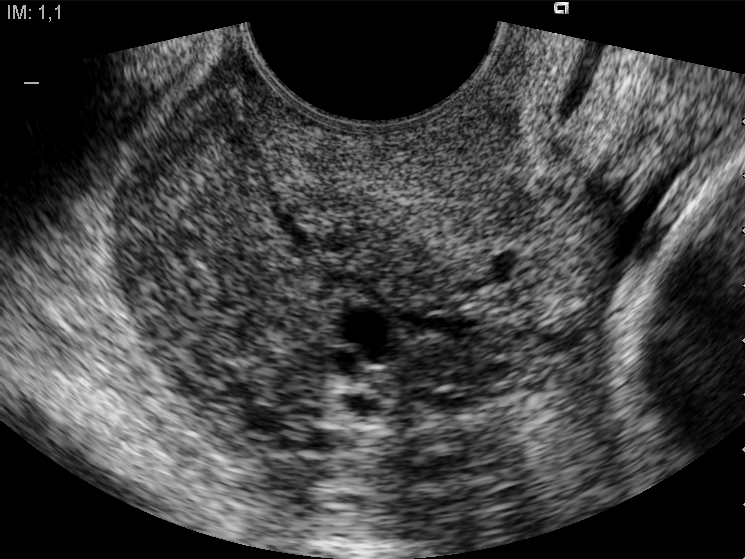}
  \caption{Maternal cervix}
\end{subfigure}\hfil 
\begin{subfigure}{0.3\textwidth}
  \includegraphics[width=2.1in,height=1.4in]{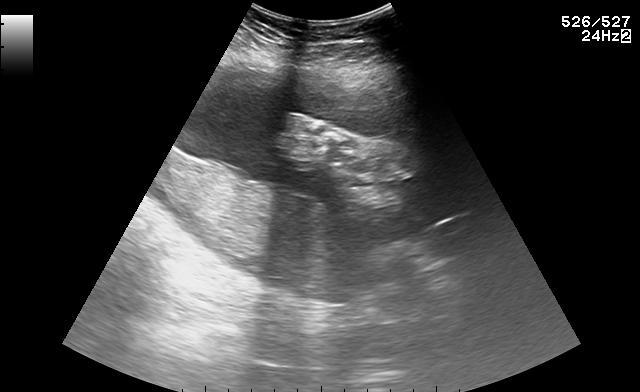}
  \caption{Other}
\end{subfigure}
\caption{Examples of real images in the FETAL\_PLANES\_DB dataset.}
\label{fig:example_images_fetal_planes_db_dataset}
\end{figure}

\subsection{Synthetic Image Generation} 

We generate synthetic images in two rounds. In each round, for each of the six classes, we generate $5000$ synthetic images, yielding a total of $30000$ synthetic images. Therefore, in total, we have $60000$ synthetic images. These images are generated with the help of the diffusion model, which conditions the image generation process on the class label, ensuring that each set of $10000$ images corresponds to a particular class. The generated images exhibit characteristics that are consistent with the anatomical features of fetal ultrasound images, as guided by the class-specific classifier.

\begin{table}[h]
\centering
\begin{tabular}{ c | c | c | c } 
\toprule
Class name & \# synthetic images  & \# synthetic images of Round $1$ & \# synthetic images of Round $2$ \\
\midrule
Fetal abdomen & $10000$ & $5000$ & $5000$ \\
\midrule
Fetal brain & $10000$ & $5000$ & $5000$ \\ 
\midrule
Fetal femur & $10000$ & $5000$ & $5000$ \\
\midrule
Fetal thorax  & $10000$ & $5000$ & $5000$ \\ 
\midrule
Maternal cervix  & $10000$ & $5000$ & $5000$ \\
\midrule
Other  & $10000$ & $5000$ & $5000$ \\
\midrule
\midrule
Total & $60000$ & $30000$ & $30000$ \\
\bottomrule
\end{tabular}
\caption{Number of synthetic images.}
\label{table:number_synthetic_images}
\end{table}

The generated images are then used to augment the real FETAL\_PLANES\_DB dataset, which contains $12700$ images. The augmented dataset is used in a downstream classification task. This synthetic data generation process not only alleviates the issue of limited labeled data but also introduces diversity in the training data, helping the model generalize better to unseen data.

\subsection{Evaluation of Generated Images: Downstream Classification Tasks}

\label{sec:downstream_classification_tasks}

First, we use $30000$ synthetic images in the first round of generation and conduct a two-stage training process. In the first stage, we pretrain the classifier on the $30000$ synthetic images generated by the classifier-guided diffusion model. These images are distributed across the six classes, and the classifier learns to distinguish between them. The pretraining process helps the classifier to better understand the data distribution and feature representations specific to each class. In the second stage, the pretrained classifier is fine-tuned using the real images from the FETAL\_PLANES\_DB dataset. The fine-tuning step ensures that the classifier adjusts to the true distribution of real ultrasound images and improves its generalization ability. This two-stage approach helps overcome the data scarcity problem by leveraging synthetic images to boost performance on a relatively small set of real data. After training, the classifier’s performance is evaluated using standard classification metrics, such as accuracy, and confusion matrix. The results are compared with classifiers trained solely on real images to assess the impact of the synthetic data. Finally, we perform similar procedure as above using the whole $60000$ synthetic images from the two rounds of generation. A second round of image generation was conducted to investigate whether increasing the number of synthetic images would lead to further improvements in classification performance.

\section{EXPERIMENTS}\label{sec4}

In this section, we describe the details of the experimental setup used to evaluate the proposed method. This includes the dataset acquisition and preparation process, data preprocessing, the architecture of the models used, and the techniques employed for image generation and classification. We also provide details on the evaluation metrics used to assess the quality of the generated images and the classification performance. All experiments in the original submission were conducted using a single Quadro RTX 5000 GPU with 16GB memory. Most experiments in the revised version were conducted using 8 A100 GPUs with 40GB memory.

\subsection{Dataset Acquisition and Preparation}

The FETAL\_PLANES\_DB is a publicly available maternal-fetal ultrasound image dataset \footnote{https://zenodo.org/records/3904280}. As shown in Table~\ref{table:fetal_planes_db_dataset}, the  FETAL\_PLANES\_DB dataset was randomly split into an $80/20$ training and test set, resulting in $9923$ images for training and $2477$ images for testing. The detailed training and testing dataset splitting ratio for each class can also be found in Table~\ref{table:fetal_planes_db_dataset}. This common split ensures a representative sample of the data is used to train the model while reserving a separate set of images to evaluate its generalization ability on unseen data. The $80\%$ portion (training set) is used to train the model. The model learns patterns and relationships within the data to make accurate predictions on new data. The $20\%$ portion (test set) was used to assess the model's performance on unseen data. This helped identify how well the model generalized to real-world scenarios beyond the training data. By splitting the data in this way, a balance between training the model effectively could be achieved and ensuring it performed well on data it has not encountered before.

The FETAL\_PLANES\_DB dataset contains a limited number of images per class, making it challenging to train robust deep learning models directly on the real data. To address this issue, we used the classifier-guided diffusion model to generate synthetic images, thereby augmenting the dataset and enhancing the generalization ability of the model.

\subsection{Data Preprocessing}

Before training the models, several preprocessing steps on the real images were performed. First, all images were normalized to have pixel values in the range $[0, 1]$. This ensures that the neural networks can learn efficiently by reducing the risk of vanishing or exploding gradients. Second, the ultrasound images in the dataset were resized to a uniform resolution of $128 \times 128$ pixels. This size was chosen to maintain sufficient detail in the images while reducing computational overhead. Each image in the dataset was assigned one of the six class labels corresponding to different fetal anatomical planes. The synthetic images generated by the diffusion model were also labelled according to their respective class. We also applied traditional data augmentation techniques to  real images. Random rotations of the images between $-30^{\circ}$ and $30^{\circ}$ were employed to simulate different scanning angles. Horizontal and vertical flippings were used to account for different fetal orientations in the ultrasound.

\subsection{Model Architectures}
\label{sec:model_architectures}

For this study, a two-part architecture was used: a diffusion model for image generation and five different classifiers for downstream classification tasks. The diffusion model was based on a state-of-the-art architecture for image generation, specifically a classifier-guided diffusion model \citep{dhariwal2021diffusion}. The model consists of an encoder-decoder architecture with several layers of convolutional and residual blocks. The diffusion process introduces random noise to the images and learns to reverse this process to generate clean, realistic images. During training, the model was conditioned on class labels to ensure that the generated images belong to the correct class. For the downstream classification tasks, we used five different classifiers to assess the performance of the synthetic images generated by the diffusion model. These classifiers represent a range of architectures, from traditional CNNs to modern transformer-based models. The details of five classifiers are as follows.
\begin{itemize}
    \item ResNet50 \citep{he2016deep}: A deep convolutional neural network with residual connections that enable it to learn deep representations while avoiding the vanishing gradient problem.
    \item DenseNet169 \citep{huang2017densely}: A variant of CNNs where each layer is connected to every other layer in a dense fashion, improving feature reuse and gradient flow.
    \item ViT\_b\_32 (Vision Transformer) \citep{dosovitskiy2020image}: A transformer-based model that splits images into patches and uses self-attention mechanisms to learn global image representations. The ``b\_32'' refers to the base version of ViT with input patch size of $32$.
    \item Swin\_t (Swin Transformer) \citep{liu2021swin}: A hierarchical transformer model designed for vision tasks. It uses shifted windowing mechanisms for efficient computation and local-global feature learning.
    \item MedMamba \citep{yue2024medmamba}: A medical-specific model that combines transformers and CNNs tailored for medical image analysis. MedMamba uses medical image priors to improve its performance on healthcare datasets.
\end{itemize}
The goal is to compare how well each of these models performs when trained on real images, synthetic images, and augmented data using the proposed method. For training the above classifiers, we used the Adam optimizer \citep{kingma2014adam}  with learning rate $0.0001$ and cross-entropy loss function. Each configuration was trained for $200$ epochs with batch size $32$, and we saved the best training and test accuracy during the training procedure. For Pretraining + Fine-Tuning, after pretraining, the checkpoint was fine-tuned on real data saved from pretraining with synthetic data, with the same setup for the optimizer and loss function.

\subsection{Diffusion Model Training and Image Generation}

During training, we used $9923$ real images from the FETAL\_PLANES\_DB dataset to learn the image distribution, as shown in Table~\ref{table:fetal_planes_db_dataset}. The diffusion model was trained to generate images that match the characteristics of the real ultrasound images in each class. Following \citet{dhariwal2021diffusion}, we used mean-squared error loss between the true noise and the predicted noise to train the diffusion model, and used cross-entropy loss to train a half U-Net to provide classifier-guided information to the diffusion model. After training, the model can generate synthetic images by conditioning on the class label. As shown in Table~\ref{table:number_synthetic_images}, in each round, for each of the six classes, we generated $5000$ synthetic images, resulting in a total of $30000$ synthetic images.

\begin{figure}[h]
    \centering 
\begin{subfigure}{0.25\textwidth}
  \includegraphics[width=1.7in]{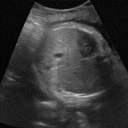}
  \caption{Fetal abdomen}
\end{subfigure}\hfil 
\begin{subfigure}{0.25\textwidth}
  \includegraphics[width=1.7in]{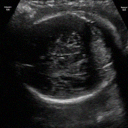}
  \caption{Fetal brain}
\end{subfigure}\hfil 
\begin{subfigure}{0.25\textwidth}
  \includegraphics[width=1.7in]{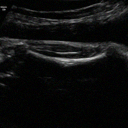}
  \caption{Fetal femur}
\end{subfigure}

\medskip
\begin{subfigure}{0.25\textwidth}
  \includegraphics[width=1.7in]{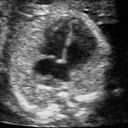}
  \caption{Fetal thorax}
\end{subfigure}\hfil 
\begin{subfigure}{0.25\textwidth}
  \includegraphics[width=1.7in]{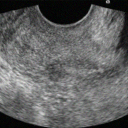}
  \caption{Maternal cervix}
\end{subfigure}\hfil 
\begin{subfigure}{0.25\textwidth}
  \includegraphics[width=1.7in]{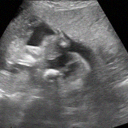}
  \caption{Other}
\end{subfigure}
\caption{Examples of synthetic images generated by the trained diffusion model.}
\label{fig:example_synthetic_images}
\end{figure}

\section{Results}
\label{sec5}

In this section, we present the experimental results obtained from our proposed method. We evaluated both the qualitative and quantitative performance of the generated images and the classifiers trained on real, synthetic, and augmented data.

\subsection{Quality of Synthetic Images}

\subsubsection{Visualization}

The generated images were visually assessed by examining a sample of synthetic images from each of the six classes. The synthetic images were compared with the real images from the FETAL\_PLANES\_DB dataset in Figure~\ref{fig:example_images_fetal_planes_db_dataset} to evaluate their realism and similarity. Example images from the diffusion model are shown in Figure~\ref{fig:example_synthetic_images}, where we observe that the synthetic images closely resemble the real images in terms of texture, structure, and class-specific features. This addition emphasizes the visual quality of the generated images, which was a key aspect of evaluating the success of diffusion models in this context.

\subsubsection{FID Score}

To evaluate the quality and diversity of the generated images, we used the Fréchet Inception Distance (FID). The FID compares the statistics of generated images with those of real images, providing a measure of how similar the two sets are. A lower FID score indicates higher similarity and better quality. The FID is calculated using the features extracted by the Inception-v3 model.

In our experiment, we classified the generated images into two classes: ``good'' and ``bad''. Initially, $5000$ ``good'' and ``bad'' images in total were manually selected. Then, a binary classifier was trained on this dataset to classify the remaining generated images. We then calculated the FID scores for both the ``good'' and ``bad'' image sets, comparing them against the real images. As shown in Table \ref{table:fid_values_good_bad}, the ``good'' images do have better quality than the ``bad'' images.

\begin{table}[h]
\centering
\begin{tabular}{ c | c | c } 
\toprule
Class name &  ``good'' images  & ``bad'' images \\
\midrule
Fetal abdomen & $148.94$ & $179.24$ \\
\midrule
Fetal brain & $118.51$ & $145.33$\\ 
\midrule
Fetal femur & $100.76$ & $130.69$ \\
\midrule
Fetal thorax  & $123.29$ & $146.91$ \\ 
\midrule
Maternal cervix  & $122.16$ & $239.34$ \\
\midrule
Other  & $90.62$ & $149.31$ \\
\bottomrule
\end{tabular}
\caption{FID values of ``good'' and ``bad'' generated images.}
\label{table:fid_values_good_bad}
\end{table}

\subsubsection{Can All Generated Images Be Used in Downstream Classification Tasks?}

Moreover, for the downstream classification task, we leveraged both the ``good'' and ``bad'' generated images, finding that both categories contributed positively to improving classifier performance, as shown in Table \ref{table:classifier_accuracy_good_bad}. Notably, however, the inclusion of ``good'' generated images yielded more improvement compared to the ``bad'' images, indicating their higher utility in enhancing the model's discriminative capabilities. On the other hand, we note that the difference in performance improvement between the two categories was not drastic, suggesting that all generated images, regardless of perceived quality, can be beneficial for downstream classification tasks. 

\begin{table}[h]
\centering
\begin{tabular}{ c | c | c } 
\toprule
Downstream classifier &  ``good'' images  & ``bad'' images \\
\midrule
Vit\_b\_32 & $88.7\%$ & $88.3\%$ \\
\midrule
Swin\_t & $92.5\%$ & $91.5\%$\\ 
\midrule
ResNet50 & $93.2\%$ & $93.2\%$ \\
\midrule
DenseNet169  & $93.3\%$ & $93.2\%$ \\ 
\midrule
MedMamba  & $93.3\%$ & $92.2\%$ \\
\bottomrule
\end{tabular}
\caption{Downstream classification accuracy of using ``good'' and ``bad'' generated images for pre-training.}
\label{table:classifier_accuracy_good_bad}
\end{table}

\subsection{Downstream Classification Results}

\subsubsection{Classification Accuracy}

The baseline for comparison is Real Data Only, which means training classifiers only on real images from the FETAL\_PLANES\_DB dataset ($9923$ images, see Table~\ref{table:fetal_planes_db_dataset}). Synthetic Data Only means classifiers trained solely on synthetic images generated by the classifier-guided diffusion model. Our method is Pretraining + Fine-Tuning, which means classifiers pretrained on synthetic images and fine-tuned on real images, as proposed and described in Section~\ref{sec:downstream_classification_tasks}.

\begin{table}[h]
\hspace{-10pt}
\centering
\begin{tabular}{ c|c|c } 
\toprule
 & Training accuracy & Test accuracy \\
\midrule
Real Data Only ($9923$ real images) & $81.6\%$ & $75.5\%$ \\ 
\midrule
Synthetic Data Only ($30000$ synthetic images) & $92.1\%$ & $83.6\%$ \\ 
\midrule
Pretraining ($30000$ synthetic images) + Fine-Tuning ($9923$ real images) & $94.2\%$ & $89.6\%$ \\ 
\midrule
\midrule
Mixed Data ($30000$ synthetic images + $9923$ real images) & $91.5\%$ & $88.3\%$ \\
\midrule
\midrule
Synthetic Data Only ($60000$ synthetic images) & $94.9\%$ & $88.4\%$ \\ 
\midrule
Pretraining ($60000$ synthetic images) + Fine-Tuning ($9923$ real images) & $96.2\%$ & $90.8\%$ \\ 
\bottomrule
\end{tabular}
\caption{Classification Performance of Vit\_b\_32: Training and Test Accuracy.}
\label{table:results_vit_b_32}
\end{table}

Table~\ref{table:results_vit_b_32} presents the results of training and test accuracy, using ViT\_b\_32 (Vision Transformer) \citep{dosovitskiy2020image} as the base classifier model, and Table~\ref{table:results_swin_t} shows the same results for Swin\_t (Swin Transformer) \citep{liu2021swin}, as mentioned in Section~\ref{sec:model_architectures}. Results of using other models, i.e., ResNet50 \citep{he2016deep}, DenseNet169 \citep{huang2017densely}, MedMamba \citep{yue2024medmamba}, can be found in Tables \ref{table:results_resnet50}, \ref{table:results_densenet169}, and \ref{table:results_medmamba}, respectively.

\begin{table}[h]
\hspace{-10pt}
\centering
\begin{tabular}{ c|c|c } 
\toprule
 & Training accuracy & Test accuracy \\
\midrule
Real Data Only ($9923$ real images) & $92.2\%$ & $89.9\%$ \\ 
\midrule
Synthetic Data Only ($30000$ synthetic images) & $95.2\%$ & $84.3\%$ \\ 
\midrule
Pretraining ($30000$ synthetic images) + Fine-Tuning ($9923$ real images) & $95.0\%$ & $91.5\%$ \\ 
\midrule
\midrule
Mixed Data ($30000$ synthetic images + $9923$ real images) & $94.6\%$ & $91.6\%$ \\
\midrule
\midrule
Synthetic Data Only ($60000$ synthetic images) & $96.5\%$ & $89.5\%$ \\ 
\midrule
Pretraining ($60000$ synthetic images) + Fine-Tuning ($9923$ real images) & $96.9\%$ & $93.1\%$ \\ 
\bottomrule
\end{tabular}
\caption{Classification Performance of Swin\_t: Training and Test Accuracy.}
\label{table:results_swin_t}
\end{table}

\begin{table}[h]
\hspace{-10pt}
\centering
\begin{tabular}{ c|c|c } 
\toprule
 & Training accuracy & Test accuracy \\
\midrule
Real Data Only ($9923$ real images) & $95.3\%$ & $92.0\%$ \\ 
\midrule
Synthetic Data Only ($30000$ synthetic images) & $96.4\%$ & $86.1\%$ \\ 
\midrule
Pretraining ($30000$ synthetic images) + Fine-Tuning ($9923$ real images) & $96.8\%$ & $92.5\%$ \\ 
\midrule
\midrule
Synthetic Data Only ($60000$ synthetic images) & $97.3\%$ & $90.4\%$ \\ 
\midrule
Pretraining ($60000$ synthetic images) + Fine-Tuning ($9923$ real images) & $97.9\%$ & $93.2\%$ \\ 
\bottomrule
\end{tabular}
\caption{Classification Performance of ResNet50: Training and Test Accuracy.}
\label{table:results_resnet50}
\end{table}

\begin{table}[h]
\hspace{-10pt}
\centering
\begin{tabular}{ c|c|c } 
\toprule
 & Training accuracy & Test accuracy \\
\midrule
Real Data Only ($9923$ real images) & $96.3\%$ & $92.0\%$ \\ 
\midrule
Synthetic Data Only ($30000$ synthetic images) & $96.7\%$ & $85.7\%$ \\ 
\midrule
Pretraining ($30000$ synthetic images) + Fine-Tuning ($9923$ real images) & $97.0\%$ & $92.7\%$ \\ 
\midrule
\midrule
Synthetic Data Only ($60000$ synthetic images) & $97.7\%$ & $91.7\%$ \\ 
\midrule
Pretraining ($60000$ synthetic images) + Fine-Tuning ($9923$ real images) & $97.5\%$ & $92.8\%$ \\ 
\bottomrule
\end{tabular}
\caption{Training and test accuracy using DenseNet169.}
\label{table:results_densenet169}
\end{table}

\begin{table}[h]
\hspace{-10pt}
\centering
\begin{tabular}{ c|c|c } 
\toprule
 & Training accuracy & Test accuracy \\
\midrule
Real Data Only ($9923$ real images) & $95.8\%$ & $91.6\%$ \\ 
\midrule
Synthetic Data Only ($30000$ synthetic images) & $96.9\%$ & $84.1\%$ \\ 
\midrule
Pretraining ($30000$ synthetic images) + Fine-Tuning ($9923$ real images) & $96.7\%$ & $92.1\%$ \\ 
\midrule
\midrule
Synthetic Data Only ($60000$ synthetic images) & $97.8\%$ & $90.6\%$ \\ 
\midrule
Pretraining ($60000$ synthetic images) + Fine-Tuning ($9923$ real images) & $98.1\%$ & $92.4\%$ \\  
\bottomrule
\end{tabular}
\caption{Classification Performance of MedMamba: Training and Test Accuracy.}
\label{table:results_medmamba}
\end{table}

\subsubsection{Confusion Matrix}

\begin{figure*}[h!tbp]
    \centering 
\begin{subfigure}{0.32\textwidth}
  \includegraphics[width=\linewidth]{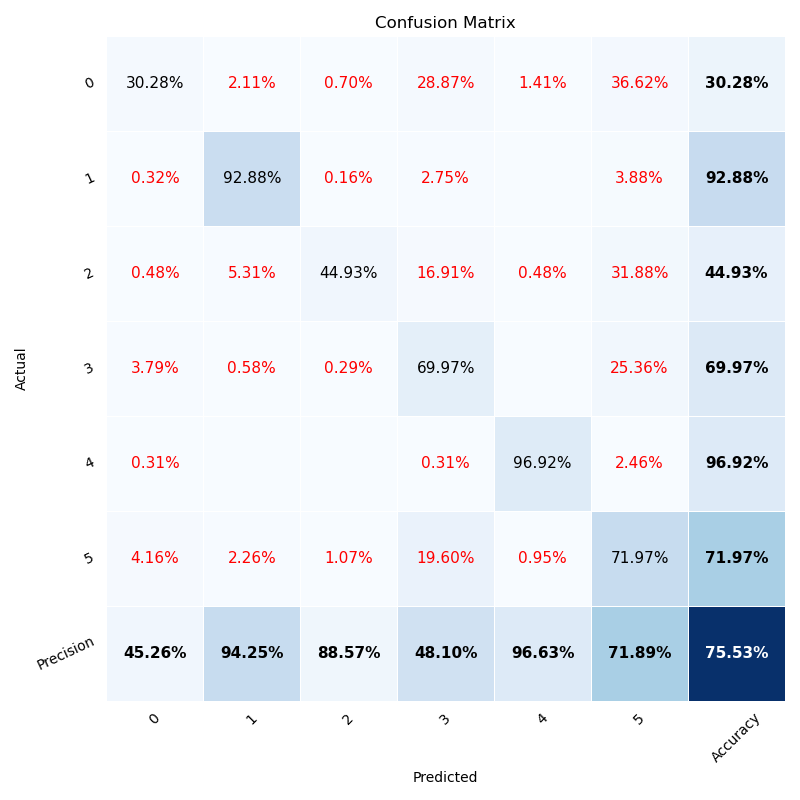}
  \caption{Real Data Only ($9923$ real images)}
  \label{fig:Vit_b_confusion_matrix_real}
\end{subfigure}\hfil 
\begin{subfigure}{0.32\textwidth}
  \includegraphics[width=\linewidth]{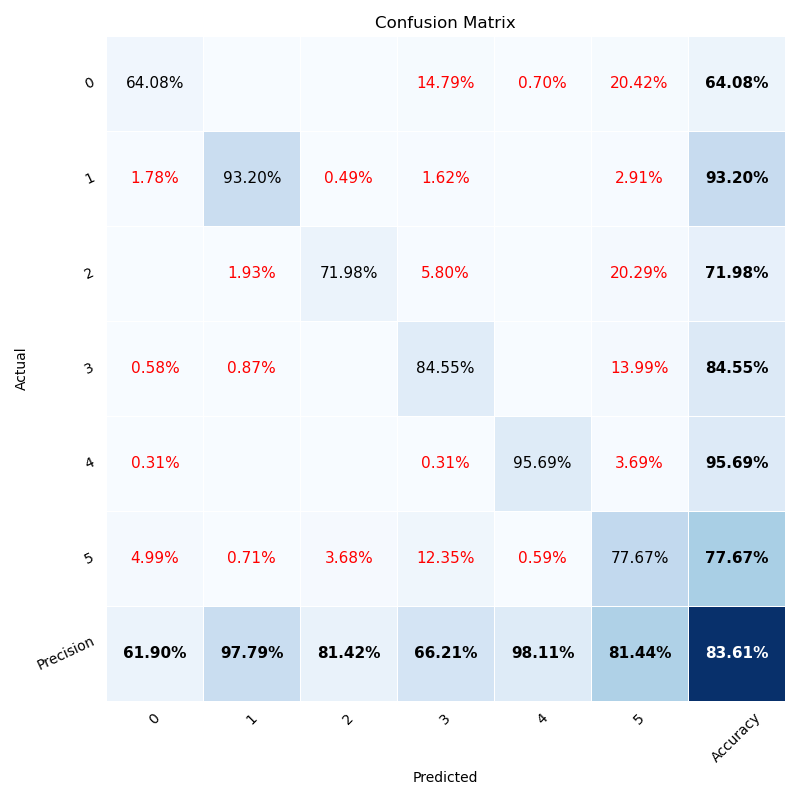}
  \caption{Synthetic Data Only ($30000$ synthetic images)}
  \label{fig:Vit_b_confusion_matrix_synthetic}
\end{subfigure}\hfil 
\begin{subfigure}{0.32\textwidth}
  \includegraphics[width=\linewidth]{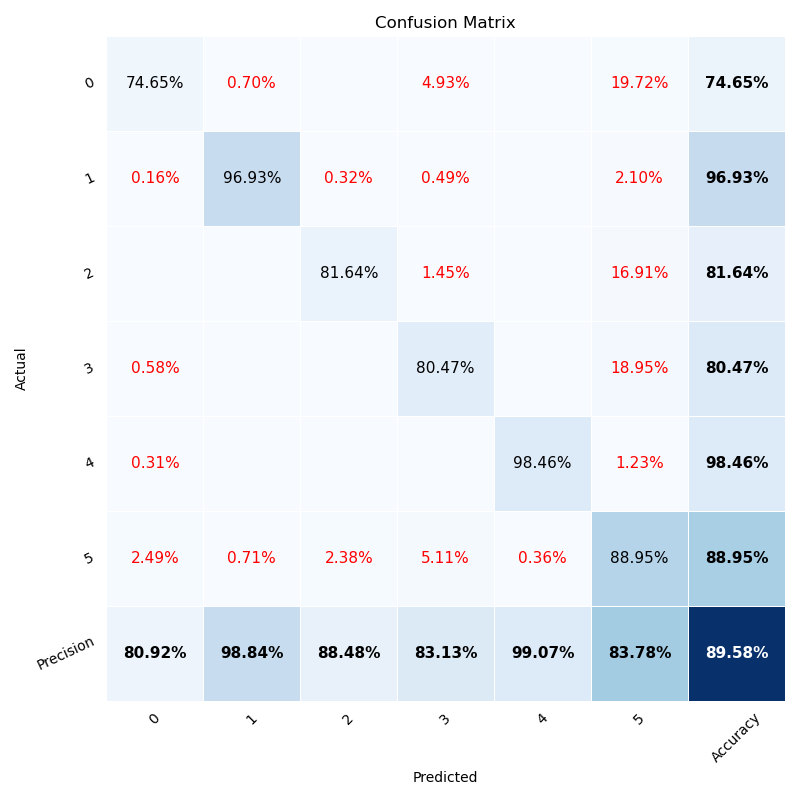}
  \caption{Pretraining ($30000$ synthetic images) + Fine-Tuning ($9923$ real images)}
  \label{fig:Vit_b_confusion_matrix_finetune}
\end{subfigure}

\medskip
\begin{subfigure}{0.32\textwidth}
  \includegraphics[width=\linewidth]{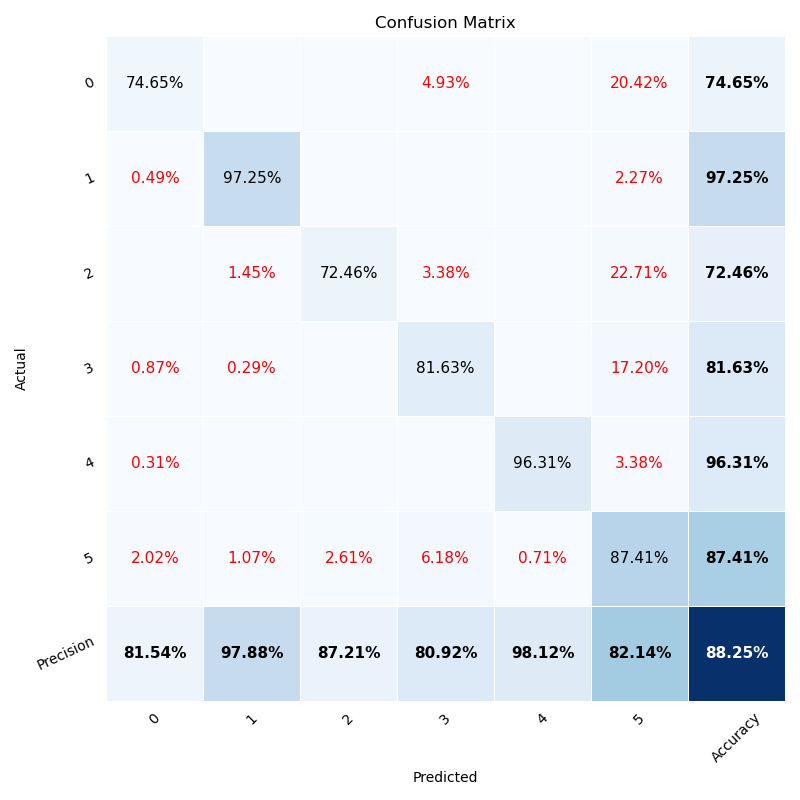}
  \caption{Mixed Data ($30000$ synthetic images + $9923$ real images)}
  \label{fig:Vit_b_confusion_matrix__mixed}
\end{subfigure}\hfil
\begin{subfigure}{0.32\textwidth}
  \includegraphics[width=\linewidth]{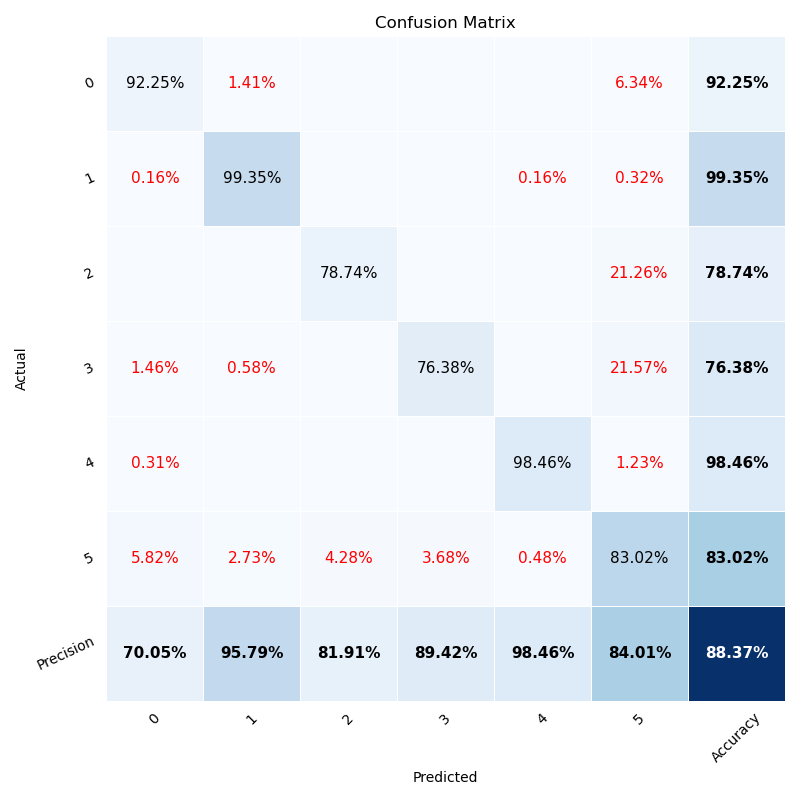}
  \caption{Synthetic Data Only ($60000$ synthetic images)}
  \label{fig:Vit_b_confusion_matrix_all_synthetic}
\end{subfigure}\hfil 
\begin{subfigure}{0.32\textwidth}
  \includegraphics[width=\linewidth]{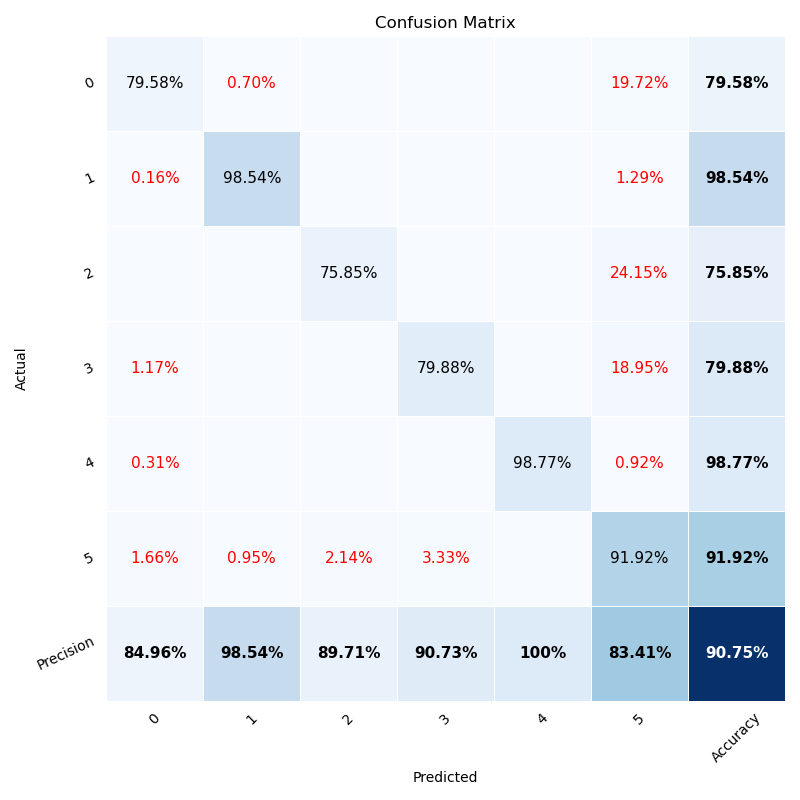}
  \caption{Pretraining ($60000$ synthetic images) + Fine-Tuning ($9923$ real images)}  \label{fig:Vit_b_confusion_matrix_all_finetune}
\end{subfigure}
\caption{Confusion matrices using Vit\_b\_32.}
\label{fig:Vit_b_confusion_matrix}
\vspace{-10pt}
\end{figure*}

To gain deeper insights into the classification performance beyond overall accuracy, we analyzed the confusion matrices for each method. The confusion matrix provides a detailed breakdown of the classifier's performance by revealing the number of true positives, true negatives, false positives, and false negatives for each class. This analysis allows us to identify specific areas where the classifier excels and where it struggles, such as class-specific biases or difficulties in distinguishing between certain classes. By examining the confusion matrices, we can pinpoint areas for improvement and refine the model accordingly.

\begin{figure}[h]
    \centering 
\begin{subfigure}{0.32\textwidth}
  \includegraphics[width=\linewidth]{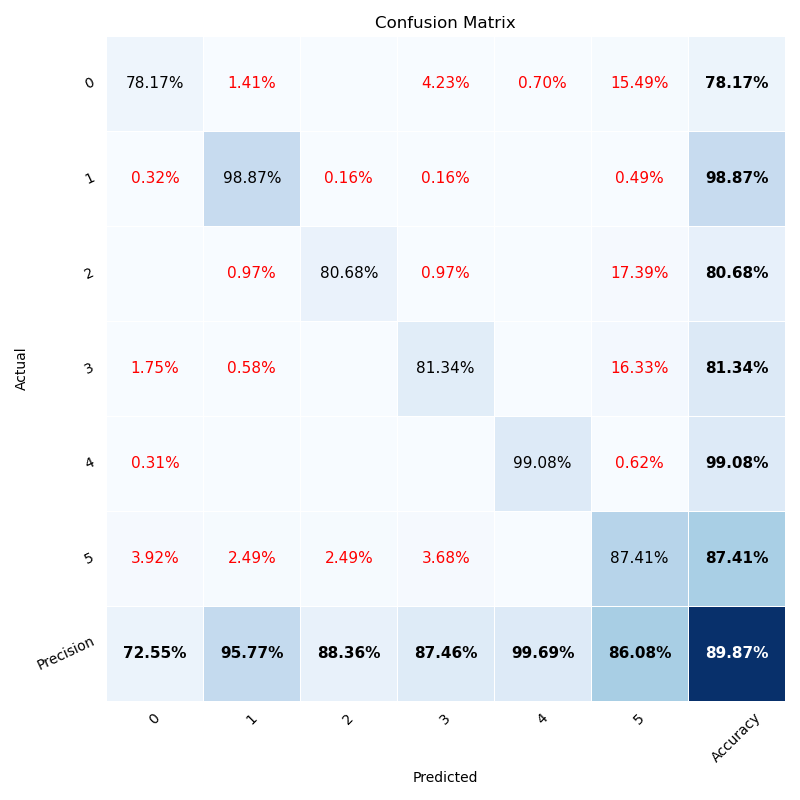}
  \caption{Real Data Only ($9923$ real images)}
  \label{fig:Swin_t_confusion_matrix_real}
\end{subfigure}\hfil 
\begin{subfigure}{0.32\textwidth}
  \includegraphics[width=\linewidth]{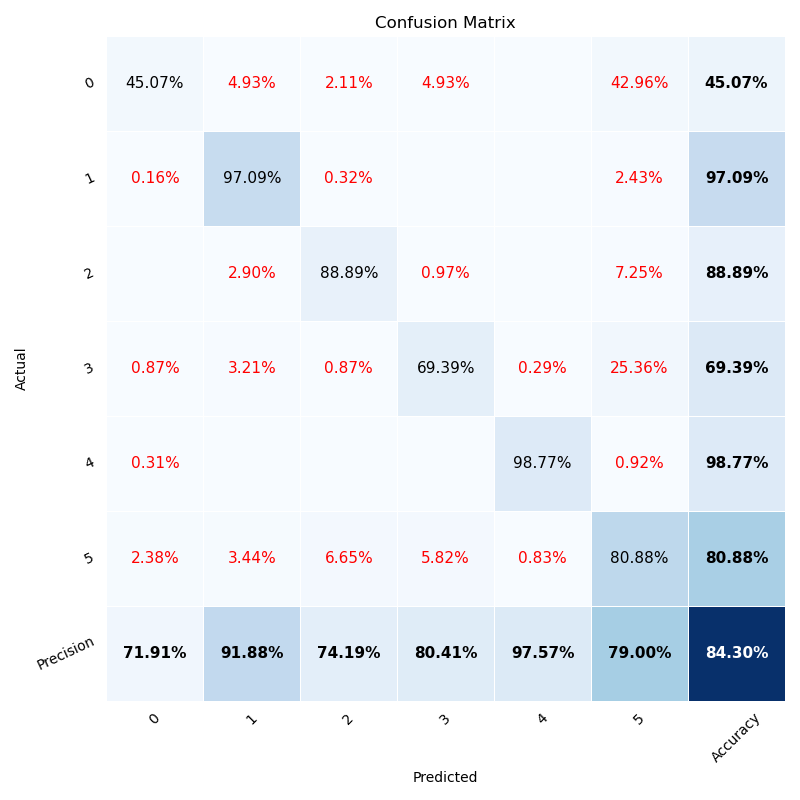}
  \caption{Synthetic Data Only ($30000$ synthetic images)}
  \label{fig:Swin_t_confusion_matrix_synthetic}
\end{subfigure}\hfil 
\begin{subfigure}{0.32\textwidth}
  \includegraphics[width=\linewidth]{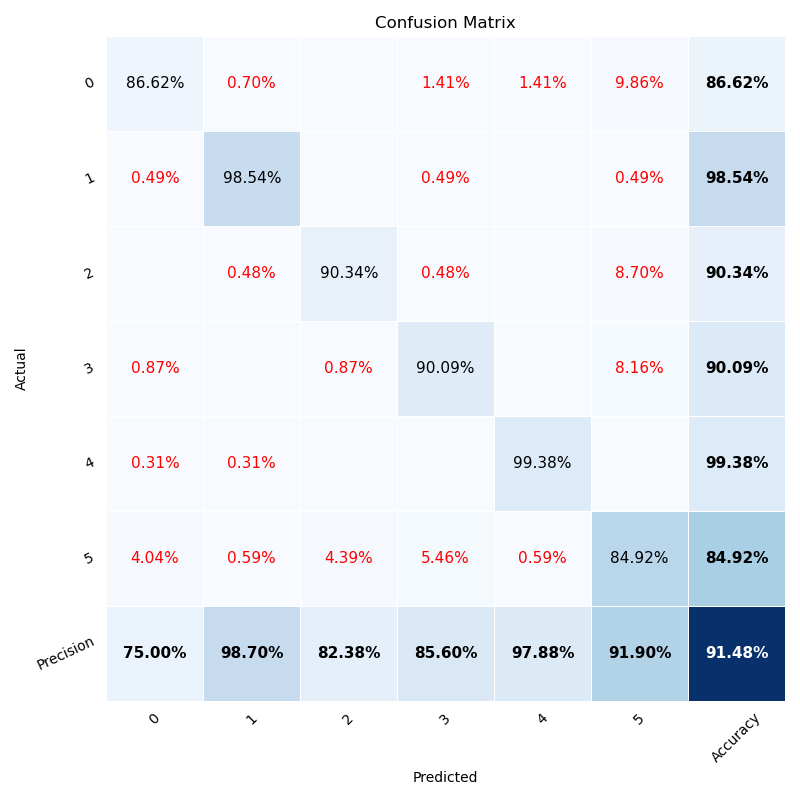}
  \caption{Pretraining ($30000$ synthetic images) + Fine-Tuning ($9923$ real images)}
  \label{fig:Swin_t_confusion_matrix_finetune}
\end{subfigure}

\medskip
\begin{subfigure}{0.32\textwidth}
  \includegraphics[width=\linewidth]{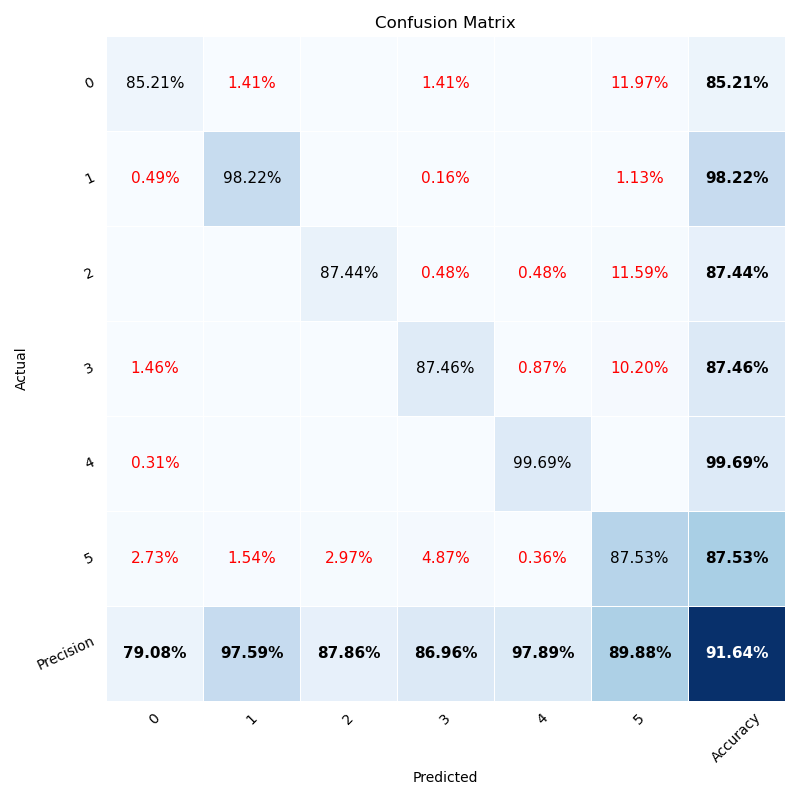}
  \caption{Mixed Data ($30000$ synthetic images + $9923$ real images)}
  \label{fig:Swin_t_confusion_matrix__mixed}
\end{subfigure}\hfil
\begin{subfigure}{0.32\textwidth}
  \includegraphics[width=\linewidth]{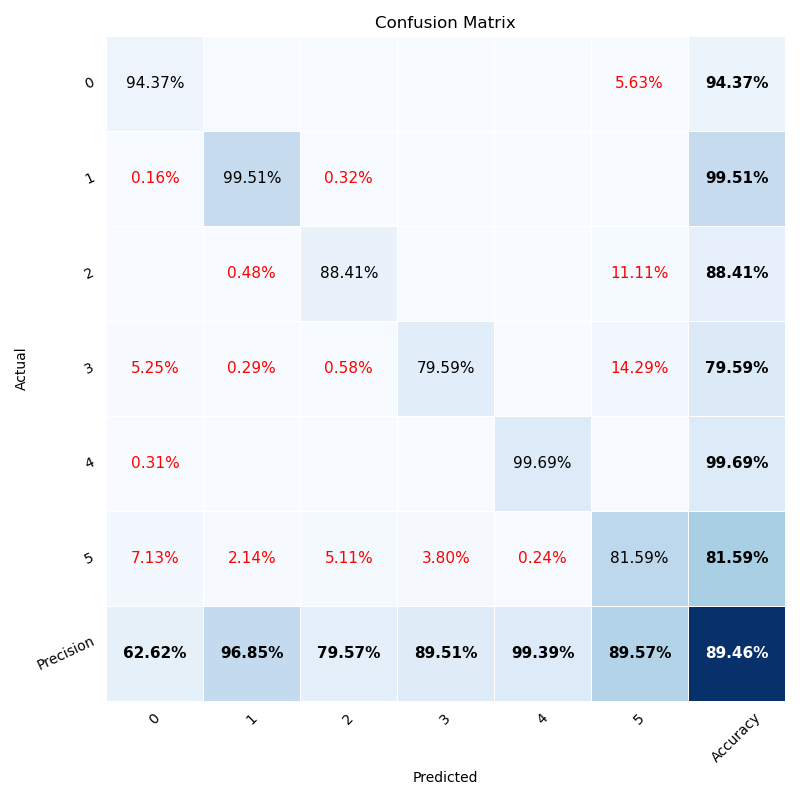}
  \caption{Synthetic Data Only ($60000$ synthetic images)}
  \label{fig:Swin_t_confusion_matrix_all_synthetic}
\end{subfigure}\hfil 
\begin{subfigure}{0.32\textwidth}
  \includegraphics[width=\linewidth]{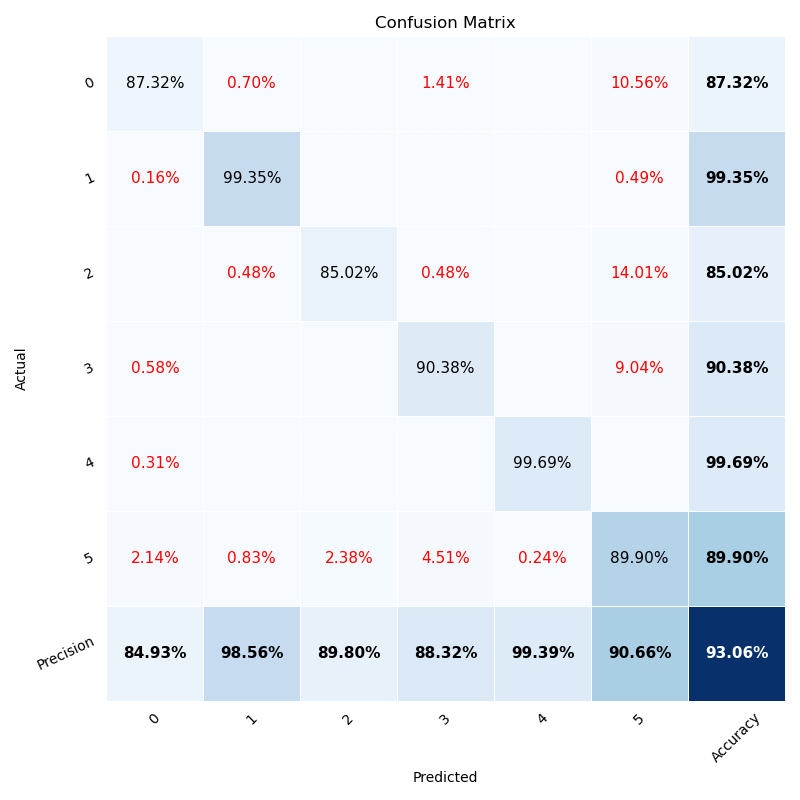}
  \caption{Pretraining ($60000$ synthetic images) + Fine-Tuning ($9923$ real images)}  \label{fig:Swin_t_confusion_matrix_all_finetune}
\end{subfigure}
\caption{Confusion matrices using Swin\_t.}
\label{fig:Swin_t_confusion_matrix}
\vspace{-10pt}
\end{figure}

According to Figure~\ref{fig:Vit_b_confusion_matrix}(a), training classifiers exclusively on real-world datasets often encounters difficulties when dealing with classes containing a limited number of images. This data scarcity leads to low test accuracy, particularly for underrepresented classes. For example, for Fetal abdomen and Fetal Femur classes, where the number of training images were $569$ and $833$ (Table~\ref{table:fetal_planes_db_dataset}), the test accuracies are only $30.28\%$ and $44.93\%$, respectively. By initially pretraining the classifier on a large dataset of synthetically generated data, its performance was significantly enhanced. The proposed two-stage process effectively addresses the issue of imbalanced classification, which is a direct consequence of data scarcity. Pretraining on synthetic data provides a more robust foundation, enabling the classifier to better recognize patterns and improve accuracy, especially for those classes with limited real-world examples. For the same two classes, i.e., Fetal abdomen and Fetal Femur classes, the test accuracy was significantly improved, achieving $74.65\%$ and $81.64\%$, respectively. Similar conclusions regarding the performance of Swin\_t can be drawn from the confusion matrices presented in Figure~\ref{fig:Swin_t_confusion_matrix}.

\subsubsection{Precision, Recall, F1 Score}

Beyond overall accuracy, we also performed a comprehensive evaluation of each individual classifier and the majority vote ensemble by calculating precision, recall, and F1-score for each of the six fetal plane classes. These metrics provide a more granular understanding of the classification performance, highlighting the ability of the models to correctly identify positive instances (precision), capture all actual positive instances (recall), and the harmonic mean of these two (F1-score). The detailed results for these metrics across all individual classifiers and the final majority vote ensemble are presented in Table~\ref{table:results_other_metrics}. These metrics were calculated using the ``weighted'' average as implemented in the scikit-learn library to account for potential class imbalances.

\begin{table}[h]
\hspace{-10pt}
\centering
\begin{tabular}{ c|c|c|c|c } 
\toprule
Classifiers & Test accuracy & Precision & Recall & F1 score \\
\midrule
ResNet50 & $93.94\%$ & $93.98\%$ & $93.94\%$ & $93.96\%$ \\ 
\midrule
DenseNet169 & $94.07\%$ & $94.08\%$ & $94.07\%$ & $94.06\%$ \\ 
\midrule
Swin\_t & $93.10\%$ & $93.13\%$ & $93.10\%$ & $93.10\%$ \\ 
\midrule
MedMamba & $93.50\%$ & $93.54\%$ & $93.50\%$ & $93.49\%$ \\ 
\midrule
Vit\_b & $90.15\%$ & $90.15\%$ & $90.15\%$ & $90.10\%$ \\ 
\midrule
ResNet50 + DenseNet169 + MedMamba (majority vote) & $94.91 \%$ & $94.91 \%$ & $94.91 \%$ & $94.91 \%$ \\ 
\midrule
ResNet50 + MedMamba + Vit\_b (majority vote) & $94.87\%$ & $94.89 \%$ & $94.87 \%$ & $94.86 \%$ \\ 
\midrule
\midrule
All $5$ classifiers (majority vote) & $94.83\%$ & $94.86\%$ & $94.83\%$ & $94.82\%$ \\ 
\bottomrule
\end{tabular}
\caption{Precision, Recall, and F1-Score for the Final Training Strategy: Pretraining with Combined Synthetic Data and Fine-Tuning.}
\label{table:results_other_metrics}
\end{table}

\subsubsection{Majority Vote}

To further enhance the robustness and accuracy of our fetal plane classification, we employed an ensemble approach. Specifically, for each of the base classifier models (ViT\_b\_32, Swin\_t, ResNet50, DenseNet169, and MedMamba), we trained five independent classifiers using the ``Pretraining + Fine-Tuning'' strategy with different random initializations. During inference on the test set, we aggregated the predictions of these five classifiers for each model using a majority voting scheme. This ensemble technique leverages the diverse learned representations of the independently trained models to mitigate individual model biases and improve the overall classification accuracy across the six fetal plane classes. The results of this majority voting ensemble are presented in Table~\ref{table:results_majority_vote}, demonstrating a significant improvement over the individual classifier performances, achieving a best accuracy of $94.91\%$.

\begin{table}[h]
\hspace{-10pt}
\centering
\begin{tabular}{ c|c } 
\toprule
Classifiers for majority vote & Test accuracy \\
\midrule
ResNet50 + DenseNet169 + Swin\_t & $94.59\%$ \\ 
\midrule
ResNet50 + DenseNet169 + MedMamba & $\mathbf{94.91\%}$ \\ 
\midrule
ResNet50 + DenseNet169 + Vit\_b & $94.67\%$ \\ 
\midrule
ResNet50 + Swin\_t + MedMamba & $94.55\%$ \\ 
\midrule
ResNet50 + Swin\_t + Vit\_b & $93.30\%$ \\ 
\midrule
ResNet50 + MedMamba + Vit\_b & $94.87\%$ \\ 
\midrule
DenseNet169 + Swin\_t + MedMamba & $94.55\%$ \\ 
\midrule
DenseNet169 + Swin\_t + Vit\_b & $93.54\%$ \\ 
\midrule
DenseNet169 + MedMamba + Vit\_b & $94.79\%$ \\ 
\midrule
Swin\_t + MedMamba + Vit\_b & $93.46\%$ \\ 
\midrule
\midrule
ResNet50 + DenseNet169 + Swin\_t + MedMamba + Vit\_b (All) & $94.83\%$ \\ 
\midrule
\midrule
Stacked ensemble of deep learning models \citep{krishna2024standard} & $94.64\%$ \\
\midrule
Deep feature integration \citep{krishna2023automated} & $93.86\%$ \\
\bottomrule
\end{tabular}
\caption{Test accuracy with majority vote ensemble.}
\label{table:results_majority_vote}
\end{table}

\subsubsection{Comparative Methods}

It is important to note that our approach differs from the related works of fetal plane classification, by focusing mostly on demonstrating the usefulness of the diffusion-generated images, while several existing works \citep{krishna2023automated,krishna2024standard,rauf2023automated,ghabri2023transfer,krishna2024automatic,krishna2024deep}, as discussed in Section \ref{sec:existing_work_fetal_plane_classification}, focused on designing specific methods for classification. 
We implemented the leading existing methods of stacked ensemble deep learning models, achieving $94.64\%$ accuracy. This is in comparison to the $95.69\%$ performance reported in the original paper by \citet{krishna2024standard}. Since their code was not publicly available, our results are based on our own implementation, carefully following the methodology and hyperparameters outlined in their paper. We also implemented another method called deep feature integration \citep{krishna2023automated}, which achieves $93.86\%$ accuracy. This is also in comparison with the $95.5\%$ reported in their original paper. Due to their code not publicly available, our results are based on our own implementation, carefully following the methodology and hyperparameters outlined in their paper.

While considerable research has focused on designing sophisticated deep learning classifiers and ensemble methods for various image datasets, such as those detailed in recent works exploring advanced CNN architectures or hybrid approaches \citep{krishna2023automated,krishna2024standard,rauf2023automated,ghabri2023transfer,krishna2024automatic,krishna2024deep}, there remains comparatively little existing work that systematically investigates the impact of diverse data augmentation strategies on their performance. Most efforts have traditionally centered on optimizing classifier architectures or training schemes, often treating traditional data augmentation (such as rotation and flip as mentioned in Section 4.3) as a standard pre-processing step rather than a primary research focus for its intricate effects.

Given the inherent benefits of data augmentation in improving model generalization and robustness, especially in scenarios with limited real data, we expect that our approach using generated images for augmentation would significantly enhance the performance of these existing state-of-the-art classifiers.

\subsection{Ablation Study}

To further validate the efficacy of our proposed method and dissect the contribution of its individual components, we conducted a series of ablation experiments. Specifically, three ablation studies were performed to assess the impact of full fine-tuning and classifier guidance in conjunction with traditional augmentation. Additionally, we employed a majority vote ensemble to further enhance the final classification performance.

\subsubsection{Effect of Traditional Augmentations}

To further validate the efficacy of our proposed pretraining strategy using diffusion-generated synthetic images, we conducted experiments incorporating standard data augmentation techniques during the fine-tuning stage. This ensures that the performance gains observed are not simply due to the introduction of more diverse data but are robust to common image transformations.

We augmented the real images from the FETAL\_PLANES\_DB dataset during the fine-tuning phase with the following transformations:
\begin{itemize}
    \item Random rotations within $30$ degrees.
    \item Clipping: Randomly clipping pixel values to simulate variations in image contrast.
    \item Random Noise: Adding Gaussian noise to the images to simulate sensor noise.
    \item Cropping: Randomly cropping regions of the images to introduce variations in object positioning and scale.
\end{itemize}
We repeated the training and evaluation procedure using the same experimental setup as described above, with the traditional augmentation applied only during the pretraining, for fair comparison. The results, presented in Table \ref{table:trad_aug_results_all}, demonstrate the performance of all the classifiers, when these augmentations are applied.

\textbf{Analysis.} As shown in Table \ref{table:trad_aug_results_all}, even with the application of standard data augmentations, the ``Pretraining + Fine-Tuning'' method of using diffusion-generated images consistently outperforms the corresponding baseline without using synthetic images. This indicates that the pretraining on diffusion-generated synthetic images provides an advantage that is not merely redundant to the benefits of traditional augmentation. The pretraining step appears to impart a more robust and generalizable representation, which translates to improved performance even when the fine-tuning data is further augmented.

These results further underscore the value of our proposed method in leveraging synthetic data to enhance model performance in medical image analysis, even in scenarios where standard augmentation techniques are employed.

\begin{table}[h]
\hspace{-10pt}
\centering
\begin{tabular}{ c|c|c|c|c|c }
\toprule
 & ResNet50 & DenseNet169 & Vit\_b & Swin\_t & MedMamba \\
\midrule
Pretraining ($30000$ synthetic images) & $86.1\%$ & $85.7\%$ & $83.6\%$ & $84.3\%$ & $84.1\%$  \\
\midrule
Pretraining + Fine-Tuning ($9923$ real) & $92.5\%$ & $92.7\%$ & $89.6\%$ & $91.5\%$ & $92.1\%$  \\
\midrule
\midrule
Pretraining ($30000$ traditional augmentation) & $93.2\%$ & $93.5\%$  & $86.6\%$ & $92.0\%$ & $92.4\%$ \\
\midrule
Pretraining + Fine-Tuning ($9923$ real) & $93.8\%$ & $93.7\%$ & $88.2\%$ & $92.7\%$ & $93.0\%$ \\
\midrule
\midrule
Pretraining (traditional + diffusion) & $93.2\%$ & $93.3\%$ & $88.5\%$ & $92.6\%$ & $92.8\%$ \\
\midrule
Pretraining + Fine-Tuning ($9923$ real) & $93.9\%$ & $94.1\%$ & $90.1\%$ & $93.1\%$ & $93.5\%$ \\
\bottomrule
\end{tabular}
\caption{Test accuracy: Effect of Traditional Augmentation on Classifier Training.}
\label{table:trad_aug_results_all}
\end{table}

\subsubsection{Effect of Classifier Guidance}

Interestingly, our experiments revealed that disabling the classifier guidance during the training of the diffusion model led to a surprising increase in the final classification accuracy. This counter-intuitive result is evident in the comparisons presented in Table \ref{table:ablation_results_classifier_guidance_all}. We speculate that this improvement arises because the classifier used for guidance, while intended to steer the generation process, might not have been sufficiently well-trained. Consequently, it could have introduced noisy or suboptimal signals to the diffusion model, hindering its ability to generate truly beneficial synthetic data for pretraining. By removing this potentially flawed guidance, the diffusion model may have learned a more robust and generalizable representation, ultimately leading to better downstream classification performance.

\begin{table}[h]
\hspace{-10pt}
\centering
\begin{tabular}{ c|c|c|c|c|c }
\toprule
 & ResNet50 & DenseNet169 & Vit\_b & Swin\_t & MedMamba \\
\midrule
\midrule
Pretraining ($30000$ synthetic images) & $86.1\%$ & $85.7\%$ & $83.6\%$ & $84.3\%$  & $84.1\%$ \\
\midrule
Pretraining + Fine-Tuning ($9923$ real) & $92.5\%$ & $92.7\%$ & $89.6\%$ & $91.5\%$  & $92.1\%$ \\
\midrule
\midrule
Pretraining ($30000$ synthetic w/o classifier guidance)  & $87.0\%$ & $83.0\%$ & $84.1\%$ & $86.1\%$ & $87.5\%$ \\
\midrule
Pretraining + Fine-Tuning ($9923$ real images) & $93.4\%$ & $92.8\%$ & $89.8\%$ & $92.2\%$ & $91.8\%$ \\
\bottomrule
\end{tabular}
\caption{Test accuracy: Impact of classifier guidance during diffusion model training.}
\label{table:ablation_results_classifier_guidance_all}
\end{table}

\subsubsection{Effect of Full Fine-Tuning}

To assess the impact of different fine-tuning strategies, we conducted ablation experiments comparing full fine-tuning against only fine-tuning the last layer (a transfer learning approach). The results of these experiments are detailed in Table \ref{table:ablation_results_full_finetuning_all}. Our findings indicate that exclusively fine-tuning the last layer was insufficient to enhance the performance of the pretrained models. In contrast, our proposed method, which involves full fine-tuning of the entire network, yielded substantial improvements over the initial pretrained model performance.

\begin{table}[h]
\hspace{-10pt}
\centering
\begin{tabular}{ c|c|c|c|c|c }
\toprule
 & ResNet50 & DenseNet169 & Vit\_b & Swin\_t & MedMamba \\
\midrule
\midrule
Pretraining (traditional + diffusion)  & $93.2\%$ & $93.3\%$ & $88.5\%$ & $92.6\%$ & $92.8\%$ \\
\midrule
Pretraining + Fine-Tuning ($9923$ real) & $93.9\%$ & $94.1\%$ & $90.1\%$ & $93.1\%$ & $93.5\%$ \\
\midrule
\midrule
Pretraining + Last Layer Fine-Tuning ($9923$ real) & $93.6\%$ & $93.6\%$ & $89.2\%$ & $93.2\%$ & $92.8\%$ \\
\bottomrule
\end{tabular}
\caption{Test accuracy: Impact of full fine-tuning during classifier training.}
\label{table:ablation_results_full_finetuning_all}
\end{table}

\section{CONCLUSION AND FUTURE WORK}\label{sec7}

In this study, we proposed a novel approach to address the challenge of data scarcity in medical image analysis, specifically for fetal ultrasound images. By leveraging a classifier-guided diffusion model, we generated $60000$ synthetic ultrasound images from the publicly available FETAL\_PLANES\_DB dataset. These synthetic images were used to pretrain classifiers, which were then fine-tuned using real images to improve performance.

We evaluated our approach using six different classifiers: ResNet50, DenseNet169, ViT\_b\_32, Swin\_t, and MedMamba. The experimental results demonstrated that our method of combining synthetic data with real data for transfer learning significantly improved the classification accuracy when compared to training with real data alone. The experimental results on downstream classification tasks showed that the synthetic images generated by our diffusion model were highly realistic and similar to real fetal ultrasound images. Our contributions can be summarized as follows:

\begin{itemize}
    \item We generated a large-scale dataset of $60000$ synthetic fetal ultrasound images, which will be made publicly available to help advance research in medical image analysis.
    \item We demonstrated that using synthetic images for pretraining and real images for fine-tuning can improve classifier performance compared to training with real images only.
    \item We provided a comparative study using multiple classifiers and found that our approach works well across a diverse set of models, including deep CNNs and transformer-based architectures.
\end{itemize}

The generated images significantly contribute to improving the classification performance. Classifiers trained solely on synthetic data (``Synthetic Data Only'') achieve a respectable level of accuracy. However, the ``Pretraining + Fine-Tuning'' approach, where classifiers were pretrained on synthetic images and then fine-tuned on real data, demonstrates a substantial improvement in both training and test accuracy compared to the baseline of training solely on real data (``Real Data Only''). This highlights the effectiveness of using synthetic data generated by the diffusion model to enhance the robustness and generalization capabilities of the classification models.


This work highlights the potential of generative models in augmenting medical datasets and enhancing classifier performance in scenarios with limited data. However, there are several directions for future work that could further improve and expand the impact of our approach:
\begin{itemize}
    \item Improved Generative Models: While the classifier-guided diffusion model demonstrated promising results, further advancements in generative modeling could be explored. For instance, the use of contrastive learning could be explored to generate more discriminative features in the synthetic images.
    \item Multimodal Imaging: While our study focuses on ultrasound images, extending this method to other types of medical imaging, such as MRI or CT scans, could offer broader applications in clinical practice. In particular, combining synthetic data from multiple modalities may help create more robust models that can generalize across different imaging techniques.
    \item Incorporating Expert Knowledge: One potential improvement is to involve domain experts, such as radiologists and sonographers, in the image generation process. This could be done by using semi-supervised learning or active learning approaches, where radiologists label or provide feedback on the generated images. Such feedback could help the model generate even more realistic and clinically relevant images.
    \item Longitudinal and Temporal Data: A key challenge in prenatal ultrasound imaging is capturing changes over time. In future work, we could explore temporal models that take into account the dynamic nature of fetal development. This would enable the generation of synthetic images that reflect the evolution of fetal structures over time, which is important for predictive models in maternal-fetal medicine.
\end{itemize}
In summary, this research demonstrates the value of combining synthetic data generation with real data for training medical image classifiers, providing a viable solution to data scarcity in ultrasound imaging. The proposed method holds promise for improving model performance in healthcare applications, and future research could build on this work to make synthetic data generation an integral part of medical image analysis, ultimately enhancing the quality and accessibility of healthcare.

\bibliographystyle{unsrtnat}
\bibliography{references}

\end{document}